\documentclass{article}

\usepackage{arxiv}

\usepackage[utf8]{inputenc} 
\usepackage[T1]{fontenc}    
\usepackage{hyperref}       
\usepackage{url}            
\usepackage{booktabs}       
\usepackage{amsfonts}       
\usepackage{nicefrac}       
\usepackage{microtype}      
\usepackage{lipsum}		
\usepackage{graphicx}
\usepackage{natbib}
\usepackage{multirow}
\usepackage{doi}
\usepackage{amsmath,amssymb,amsfonts}
\usepackage{xcolor}

\usepackage{hyperref}

\definecolor{newcolor}{rgb}{.8,.349,.1}
\usepackage{subcaption} 

\title{Brain Networks and Intelligence: A Graph Neural Network Based Approach to Resting State fMRI Data}

\author{\thanks{Corresponding Author: Bishal Thapaliya, bthapaliya16@gmail.com.}\, Bishal Thapaliya \textsuperscript{1, 2} \quad Esra Akbas \textsuperscript{1} \quad Jiayu Chen \textsuperscript{1, 2} \quad Raam Sapkota \textsuperscript{1, 2} \quad Bhaskar Ray \textsuperscript{1, 2} \quad \\\textbf{Pranav Suresh \textsuperscript{1, 2}} \quad \textbf{ Vince Calhoun \textsuperscript{1, 2, 3}} \quad \textbf{Jingyu Liu \textsuperscript{1,2}}\\
\textsuperscript{1}Georgia State University  \quad  \textsuperscript{2}TReNDs Center \quad \textsuperscript{3}Georgia Institute of Technology
\\}

\renewcommand{\shorttitle}{\textit{Brain ROI-aware
Graph Isomorphism Networks (BrainRGIN )} }

\begin{document}
\maketitle

\newcommand{\technique}{\textit{BrainRGIN }}
\maketitle

\begin{abstract}
Resting-state functional magnetic resonance imaging (rsfMRI) is a powerful tool for investigating the relationship between brain function and cognitive processes as it allows for the functional organization of the brain to be captured without relying on a specific task or stimuli. In this paper, we present a novel modeling architecture called BrainRGIN for predicting intelligence (fluid, crystallized and total intelligence) using graph neural networks on rsfMRI derived static functional network connectivity matrices. Extending from the existing graph convolution networks, our approach incorporates a clustering-based embedding and graph isomorphism network in the graph convolutional layer to reflect the nature of the brain sub-network organization and efficient network expression, in combination with TopK pooling and attention-based readout functions. We evaluated our proposed architecture on a large dataset, specifically the Adolescent Brain Cognitive Development Dataset, and demonstrated its effectiveness in predicting individual differences in intelligence. Our model achieved lower mean squared errors, and higher correlation scores than existing relevant graph architectures and other traditional machine learning models for all of the intelligence prediction tasks. The middle frontal gyrus exhibited a significant contribution to both fluid and crystallized intelligence, suggesting their pivotal role in these cognitive processes. Total composite scores identified a diverse set of brain regions to be relevant which underscores the complex nature of total intelligence.
\end{abstract}


\section{Introduction}

Intelligence is a complex construct that comprises various cognitive processes. Researchers typically rely on a range of cognitive tests that measure different aspects of cognition and form specific measures for intelligence, such as fluid intelligence (the ability to reason and solve problems in novel situations)(\cite{Kyllonen2017}), crystallized intelligence (the ability to use knowledge and experience to solve problems), and total intelligence (a composite measure of overall cognitive ability). There is always undamped human interest to reveal neural underpinnings of such intelligence and to predict individual intelligence differences. Although traditional MRI studies in the literature focused on structural brain measures for various phenotypes (\cite{Suresh2023, Ray2023}), rapidly growing studies have emerged to investigate the prediction of intelligence based on brain functional features (\cite{Ferguson2017,He2020,Dubois2018}). The review conducted by (\cite{Vieira2022}) further concluded that functional MRI (fMRI) has become the most employed modality for predicting intelligence, and resting-state fMRI (rs-fMRI) derived static functional connectivity (FC) was the most studied predictor. rs-fMRI measures spontaneous brain activity during rest through blood oxygen level-dependent (BOLD) signals in response to neuronal activity. FC is defined as the degree of temporal correlation between regions of the brain computed using time series of BOLD signals, and rs-fMRI FC provides a comprehensive view of the brain's intrinsic organization (\cite{Lee2012}). Indeed, FC between the default mode network and frontoparietal network have been validated to contribute to individual differences in cognitive ability (\cite{Hearne2016}). \\

While the majority of intelligence prediction approaches involve linear regression methods, some studies have applied non-linear approaches including polynomial kernel SVR (\cite{Wang2015}), and kernel ridge regression (\cite{He2020}) methods and deep neural networks (\cite{He2020, Fan2020, Li2023}). In recent years, Graph Neural Networks (GNNs) have gained great interest and have evolved rapidly for end-to-end graph learning applications. GNNs are considered the state-of-the-art deep learning methods for solving graph-structured data analysis problems, as they specify a neural network to fit into the graph structure with nodes and edges, and embed node features and edge features with structural information in the graph. Various studies have investigated the effectiveness of GNNs in different applications such as social networks, protein networks, and neurological biomarkers (\cite{Kim2020, Kazi2023, Nandakumar2021}).
Given the network-structured nature of the brain, modeling brain connectome via GNNs has been implemented. Most brain GNN studies utilize the FC graph from rs-fMRI (\cite{Ktena2018,koch2022,Wu2021, Guixiang2018}) and classify a particular phenotype of the subjects, such as gender (\cite{Salim2018,Anees2021}) or specific disease status (\cite{Anees2021,Guixiang2018}), while its predictive power for intelligence is yet to be investigated. \\

Most GNNs assume that the nodes learn embeddings in an identical way throughout the whole graph, which is problematic for brain connectome due to the sub-network nature of the brain (\cite{Parente2020}). Recently, BrainGNN (\cite{Li2021}) proposed a new GNN architecture that tackled this limitation by proposing a clustering-based embedding method in the graph convolutional layer, which allows nodes in different clusters (representing different brain networks) to learn embeddings differently. Inspired by BrainGNN, we present a novel GNN model, Brain ROI-aware Graph Isomorphism Networks (\technique), for intelligence prediction. Firstly, we make use of a graph isomorphism network (GIN) (\cite{GIN2018})  to improve the expressive power of GNNs, which is designed to approximate the power of the Weisfeiler-Lehman (WL) graph isomorphism test.  Similar to BrainGNN, we also tackle the limitation of node identical learning mechanism by introducing cluster representation of the region of interest (ROI). 
By combining these two architectures, our model can effectively capture both local and global relationships between brain regions. Furthermore, we validate various aggregation and read-out functions including attention-based readout methods. As far as we are aware, this is the first study that uses graph neural networks to uncover the patterns of the brain for intelligence prediction using resting state fMRI data. We evaluate the performance of our proposed model on a large dataset and demonstrate its effectiveness in predicting individual differences in intelligence.

\begin{equation}
h_i^{(l)} = MLP^{(l)} \left( \left( 1 + \epsilon^{(l)} \right) h_i^{(l-1)} + \sum_{j \in \mathcal N^{(l)}(i)} h_j^{(l-1)} \right )
\end{equation}

where $\epsilon^{(l)}$ is a trainable parameter that helps to avoid the over-smoothing problem, and $MLP^{(l)}$ is a 2-layered multi-layer perception to update the final feature representation for the $i$-th node in the $l$-th layer.
The output of the graph is then computed as:

\begin{equation}
h_G = READOUT \left( { h_i^{(l)} }_{v \in G} \right)
\end{equation}

where $h_G$ is the final representation of the graph and $READOUT$ is a permutation invariant function that aggregates the node representations to produce a single output vector. GINs have been shown to outperform other state-of-the-art GNNs on various graph classification tasks, demonstrating their power and effectiveness in learning representations of graph-structured data (\cite{GIN2018}).

\subsection{Comparision with existing deep learning based architectures for Brain Network Analysis}
In recent times, there has been a growing focus on extending the applicability of Graph Neural Network (GNN)-based models to the realm of brain network analysis. Notably, BrainGNN (\cite{Li2021}) innovates by crafting Region of Interest (ROI)-aware GNNs to harness functional information within the brain and employing a specialized pooling operator to identify pivotal nodes.  FBNetGen (\cite{FBNetGen}) delves into the trainable generation of brain networks while investigating their interpretability for downstream tasks. Graph Transformer \cite{GraphTransformer} uses static FC as features and applies multi-head attention with residual connections and normalization blocks to perform downstream tasks. A much more recent model, Brain Network Transformer (BNT) (\cite{BrainNetworkTransformer}) effectively used transformer-scaled dot product attention and introduced an orthonormal clustering readout function that leads to cluster-aware node embeddings. On the other hand, IBGNN (\cite{IBGNN}) initializes a mask based on edges and refines it with sparse control during training to identify top connections and regions for a specific task, aiming to create a subnetwork capable of predicting information as effectively as the original network. Unlike IGBNN which pools connections (edges) based on the sparse mask, we aim to select top salient ROIs by pooling nodes. While these models seem to be very effective, we direct our focus on designing models to harness the power of isomorphism networks during graph-based convolution, and attention-based readouts for intelligence prediction. As for the conceptualization, our model relates very closely to the BNT, which uses a clustering-based embedding approach during the readout after scaled dot product attention-based node feature updates. Our model, however, implements clustering-based node feature updates during the isomorphism-based graph convolution part, followed by scaled dot product attention for the readout function. The complete idea is further elucidated in Section \ref{Proposed_Architecture}.

\section {Proposed Architecture}
\label {Proposed_Architecture}
 \begin{figure*}[ht]
\includegraphics[width=\linewidth]{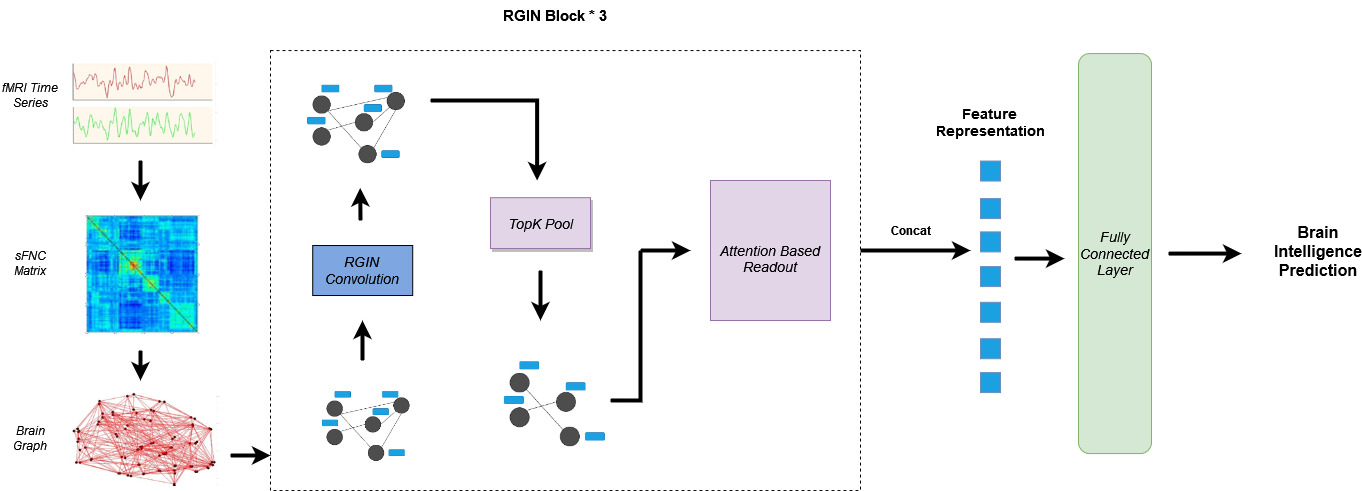}
\caption{  Overall Architecture of \technique.  The static FNC matrix is extracted from a resting state fMRI time series data. Three blocks of BrainRGIN are used with attention-based readout functions followed by a fully connected layer for prediction. }
\label{BrainRGIN_Architecture}
\end{figure*}

The architecture presented in Fig. \ref{BrainRGIN_Architecture} depicts the overall working mechanism of the \technique. As the initial step, we create a functional connectivity graph using functional relations between ROIs of the brain. After creating the FC graph, a three-layer RGIN (ROI-aware graph isomorphism network) block is established, where each block encompasses a sequence of RGIN convolutions, succeeded by a TopK Pooling layer. 
 The output of RGIN is subject to an Attention-Based Readout block either with a Squeeze-Excitation Readout (SERO) function or a Graph Attention Readout (GARO) function. Finally, a single-layer fully connected neural network serves to compute the intelligence score.\\
 
\subsection{Functional connectivity graph creation}

The initial phase involves the processing of input from fMRI time series data to create a static FC matrix or static functional network connectivity (sFNC) matrix. The difference between FC and FNC lies in how ROIs are generated: regions selected based on an atlas lead to an FC matrix, while regions generated by independent component analyses (ICA) lead to an FNC matrix. The brain is spatially divided into N ROIs. In this study, we employed ICA ROIs resultant static FNC matrix.
 Using ROIs time series, Pearson's correlation between ROIs is used to create a FNC matrix. The resulting FNC matrix is used to construct an undirected graph denoted by the tuple \begin{math} G = (V, E) \end{math}  with an adjacency matrix $A \in R^{N*N}$.
Each ROIs represents graph nodes indexed by the set \begin{math} V = 1,..., N \end{math}. The input node features,  \begin{math} x_i\end{math}, are formed using Pearson's correlation coefficients between node i and all other nodes. An edge set \begin{math} E\end{math} represents the functional connections between ROIs with each edge  \begin{math} e_{ij} \end{math} linking two nodes \begin{math} (i,j) \in E \end{math}. 

The Adjacency matrix $A_k \in R^{N \times N}$ is formed as in Equation \ref{eq:adj}, where $e_{i,j} $  is the thresholded element of FNC matrix to achieve either fully connected or sparse graph. 

\begin{equation}
    A_{i,j}^k = \left\{
    \begin{array}{ll}
        0, & i=j \\
        e_{i,j}, & \text{otherwise}
    \end{array} \right.
\label{eq:adj}
\end{equation}

\subsubsection{RGIN Convolution }
We present RGIN convolution, an approach to effectively capture both node and edge features in the graph by merging RGCN and GIN. Similar to BrainGNN, within the context of brain function, the type of edge relation reflects the different functional clustering of brain networks. In other words, nodes in the brain are divided into functional clusters, and different brain clusters form unique types of edge relations.
Without fixing cluster assignments for ROIs, the proposed model also learns cluster formation such that a group of nodes in the same cluster works together (update together) to achieve the best performance for the task. GIN is incorporated for learning node embeddings for its better graph discriminative expression.

 In RGIN, we replace the aggregation function of GIN with that of RGCN, allowing the model to learn different mechanisms for different clusters. The motivation behind using this design is twofold: 1) GIN utilizes Multilayer Perceptrons (MLPs) to learn the injective functions to adapt to the complexity of brain connectivity. 2) RGCN is able to model the sub-network clustering nature of brain connectivity. The forward propagation function of RGIN is defined as follows:



\begin{equation}
h_i^{(l)} = \text{MLP}^{(l)} \left( (1+\epsilon^{(l)})\cdot W_i^{(l)} \cdot h_i^{(l-1)} + \sum_{j\in N^{(l)}_{(i)}} W_j^{(l)} \cdot e_{i,j}^{(l-1)} \cdot h_j^{(l-1)}\right)
\end{equation}

While in regular GNN models, including GIN and RGCN, $ W_i $ and $W_j $ are learnable weight matrices, we define them as the function with parameters subject to cluster assignment of nodes to make it ROI aware convolution layer. Then, RGIN is trained in an end-to-end fashion with a 2-layered MLP. $\epsilon^{(l)}$ here is a learnable parameter that determines the importance of node $i$ compared to its neighbors $j$. In addition, multiplying node features by the edge weights, similar to the graph convolution operations in (\cite{Gong2018}), created a greater impact for neighbors connected by stronger edges.

As given in Equation ~\ref{eq:wi},  ${W}^{(l)}_i$ is defined as a function to incorporate the position encoding $r_i$, with parameters $\theta_1$ and $\theta_2$ defined to learn the clustering specific parameter for nodes. $\text{ReLU}$ is the rectified linear unit activation function. and $\textbf{b}^{(l)}$ is the bias term.
\begin{equation}
\textbf{W}^{(l)}_i  = \theta_2^{(l)} \cdot \text{relu}(\theta_1^{(l)} \textbf{r}_i) + \textbf{b}^{(l)}
\label{eq:wi}
\end{equation} 


 In the same way as BrainGNN (\cite{Li2021}), we allow RGIN to learn different update learning parameters for different clusters conditioned on the ROIs. We represent each node's location information by a vector $\textbf{r}_i$, using one-hot encoding instead of coordinates, assuming that the ROIs are aligned in the same order for all the brain graphs. 


We assume that $\theta^{(l)}_1$ = $[\alpha^{(l}_1, ......, \alpha^{(l}_{N^{(l)}}]$, where $N^{(l)}$ is the number of ROIs in the $l^{th}$ layer, $
\alpha^{(l)}_i = [\alpha^{(l)}_{i1}, \ldots, \alpha^{(l)}_{iK(l)}] \in \mathbb{R}^{K(l)}, \forall i \in \{1, \ldots, N^{(l)}\}
$, where \(K(l)\) is the number of clusters in \(N^{(l)}\) ROIs, and $ \alpha^{(l)}_i$ is the non-negative assignment scores of ROI $i$ to clusters.  Assume \(\mathbf{\theta}_2^{(l)} = [\beta^{(l)}_1, \ldots, \beta^{(l)}_{K(l)}]\) with \(\beta^{(l)}_u \in \mathbb{R}^{d(l+1) \times d(l)}\), \(\forall u \in \{1, \ldots, K(l)\}\). $\beta^{(l)}_u$ is a basis matrix. Then, we can rewrite equation (\ref{eq:wi}) as: 
\begin{equation}
\textbf{W}^{(l)}_i = \sum_{u=1}^{K(l)} {\alpha}_{iu}^{(l)}  \beta^{(l)}_u + \textbf{b}^{(l)} 
\end{equation}


This reduces the number of learnable parameters while still allowing a separate embedding kernel for each ROI. The final forward propagation function of RGIN Convolution can be written as:
\begin{equation}
\begin{split}
h_i^{(l)} = \text{MLP}^{(l)} \Biggl( &(1+\epsilon^{(l)}) \cdot \left(\sum_{u=1}^{K(l)} \alpha_{iu}^{(l)} \beta^{(l)}_u + \textbf{b}^{(l)}\right) \cdot h_i^{(l-1)} \\
&+ \sum_{j\in N^{(l)}_{(i)}} h_j^{(l-1)} \cdot \left(\sum_{u=1}^{K(l)} \alpha_{ju}^{(l)} \beta^{(l)}_u + \textbf{b}^{(l)} \right) \cdot e_{i,j}^{(l-1)} \Biggr)
\end{split}
\end{equation}

\subsubsection{Pooling Layers}
 Via RGIN convolution, node-wise representations are generated. However, for the prediction task discussed in this paper, which requires graph-level prediction instead of node-level, these node-wise representations must be collected or pooled together. We apply ROI-aware TopK Pooling as it improves interpretability for rsfMRI data, by keeping the most indicative ROIs and removing the noisy and uninformative nodes (\cite{Li2021}), along with enforcing sparsity on the network. 
To select the most indicative ROIs, the choice of which nodes to drop is determined based on the scores of nodes obtained by projecting the node features to 1D via a learnable projection vector $\omega^{(l)} \in \mathbb{R}^{d^{(l)}}$. The pooled graph $(V^{(l+1)}, E^{(l+1)})$ is computed as follows:


\begin{enumerate}
  \item Calculate the scores of nodes $s^{(l)}$ with node feature matrix as follows:

   \begin{align}
   s^{(l)} = \frac{{H}^{(l)} w^{(l)}}{\|w^{(l)}\|_2}
   \end{align}

  \item Normalize the score $s^{(l)}$ by subtracting its mean and dividing by its standard deviation, yielding final scores $\tilde{s}^{(l)}$:

 \begin{align}
   \tilde{s}^{(l)} = \frac{(s^{(l)} - \mu(s^{(l)}))}{\sigma(s^{(l)})}
   \end{align}

  \item Find the top $k$ elements in the normalized score vector $\tilde{s}^{(l)}$ with $k$- largest values and get their indices as $i$:

 \begin{align}
   i = \text{topk}(\tilde{s}^{(l)}, k)
    \end{align}

  \item Create the pooled node features $H^{(l+1)}$ by element-wise multiplying the original node features $H^{(l)}$ with the sigmoid of the normalized scores and selecting only the elements indexed by $i$:

 \begin{align}
   H^{(l+1)} = (H^{(l)} \odot \text{sigmoid}(\tilde{s}^{(l)}))_{i:}
\end{align}

  \item Create new adjacency matrix as the new pooled graph $E^{(l+1)}$ by returning the edges between selected nodes $i$ from previous edge matrix $E^{(l)}$

   \[
   E^{(l+1)} = E^{(l)}_{i,i}
   \]
\end{enumerate}

In this description, $\|\cdot\|_2$ represents the L2 norm, $\mu$ and $\sigma$ are functions that calculate the mean and standard deviation of a vector, \text{TopK} identifies the indices of the largest $k$ elements in a vector, $\odot$ denotes element-wise multiplication, and $(\cdot)_{i,:}$ selects elements in the $i^{th}$ row and all columns. \\

\subsubsection{Readout}
The conventional READOUT function of GNN can be thought of as a fixed decoder that decodes whole-graph features from the node features with no learnable parameters. We address this issue by incorporating attention to the READOUT function. The attention here refers to the scaling coefficient across the nodes learned by the model. Therefore, we make use of attention-based readout functions called GARO and SERO (\cite{Squeeze}) as proposed in STAGIN (\cite{STAGIN}).\\

GARO function  is inspired by the key-query embedding-based attention mechanism of the Transformer (\cite{vaswani2017attention}). The key embedding $\mathbf{K} \in \mathbb{R}^{D \times N}$ is computed from node feature matrix $\mathbf{H} \in \mathbb{R}^{D \times N} $, $D$ is the dimension of node feature and $N$ is a number of nodes. The query embedding $\mathbf{q} \in \mathbb{R}^{D}$  is computed from the vector of unattended graph representation $\mathbf{H}_{\phi{mean}}$, where ${\phi{mean}}$ represents the global average of node feature matrix $H$. $zspace$ is the final attention vector.
\begin{align}
\mathbf{K} &= \mathbf{W}_{key}\mathbf{H}, \\
\mathbf{q} &= \mathbf{W}_{query}\mathbf{H}{\phi_{mean}}, \\
\mathbf{z}{space} &= \sigma\left(\frac{\mathbf{q}^{\top}\mathbf{K}}{\sqrt{D}}\right),
\end{align}

where $\mathbf{W}_{key} \in \mathbb{R}^{D \times D}$, $\mathbf{W}_{query} \in \mathbb{R}^{D \times D}$ are learnable key-query parameter matrices, $\sigma$ is the sigmoid function. The GARO function uses the attended key matrix to compute the final graph representation as follows:
\begin{align}
\mathbf{h}_{\mathrm{GARO}} &= \mathbf{H}\mathbf{z}_{space}.
\end{align}
The GARO function allows the graph neural network to effectively capture dynamic brain connectome by attending to the important nodes.\\

On the other hand, the SERO function applies the squeeze and excitation operations to enhance the relevant features. Originally, the squeeze excitation module was shown to increase the performance of CNN models (\cite{Roy2019}). As we have a fixed number of nodes similar to (\cite{STAGIN}), we applied the SERO module. It is important to note that the SERO activation does not scale the channel dimension, but the node dimension. The squeeze operation computes the global information of the node features using the global average pooling, followed by the excitation operation which models the dependencies between different nodes. Specifically, we first apply a linear transformation to the squeezed feature vector and pass it through the sigmoid function to obtain the importance scores for each nodes. These importance scores are then multiplied with the original feature vector to obtain the final readout feature vector. Given a set of node features $\mathbf{H}$ with dimensionality $D$, the SERO function outputs the  feature vector $\mathbf{h_{SERO}}$ of dimensionality $\mathbb{D'} \in \mathbb{R}^{N \times D}$ as:


\begin{align}
z &= \mathbf{H}{\phi_{mean}}\\
{z}{space} &= \sigma(\mathbf{W}_2 \delta(\mathbf{W}_1 z)) \\
\mathbf{h}_{SERO} &=\mathbf{H}\mathbf{z}_{space}.
\end{align}

where $\sigma$ is the sigmoid function, $\delta$ is the ReLU activation function, $\mathbf{W}_1 \in \mathbb{R}^{D \times D}$, $\mathbf{W}_2 \in \mathbb{R}^{N \times D}$ are learnable weight matrices. The first equation computes the global information of the node features using global average pooling, and the second equation models the channel-wise dependencies by learning the feature importance scores. The third equation combines the original feature vectors with the computed importance scores to obtain the final readout feature vector.\\

Furthermore, as the effectiveness of the mean and max element-wise pooling operation is proven (\cite{He2020}), we compare the attention-based readout function with the mean and max element-wise pooling readout method. For a network, $G$ with $l$ convolution and respective pooling layers, the output graph of the $l^{th}$ pooling block is summarized using a mean and max pooling operation element-wise on $H(l)=[h(l)_i:i=1,...,N(l)]$. The resulting vector is obtained by concatenating both the mean and max summaries. To obtain a graph-level representation, the summary vectors from each layers of RGIN block are concatenated together.\\

\subsubsection{Loss Functions}

We primarily use the SmoothL1Loss function as a criterion to train the model. SmoothL1Loss is preferred over Mean Squared Error (MSE) Loss as it provides a more robust and less sensitive measure of distance between predicted and target values, with a less steep gradient around zero that makes it less sensitive to outliers and noise, leading to better model convergence and improved performance. It is defined as:

\[
L_{\text{smoothL1}}(x, y) = 
\begin{cases}
    0.5 \cdot (x - y)^2, & \text{if } |x - y| < 1 \\
    |x - y| - 0.5, & \text{otherwise}
\end{cases}
\]

Here, \(x\) represents the predicted value and \(y\) is the target value.\\

Furthermore, we also add the Unit Loss and TopK pooling loss (\cite{Li2021}) for interpretation. Unit Loss is important because we use a learnable vector $\mathbf{\omega}$ to represent nodes, but this vector can be scaled in different ways without affecting the results. To solve this issue, we add a rule that the vector must have a fixed length of 1, and we use a loss function to ensure that this rule is followed, which is Unit Loss. The loss is formulated as:
\[
L_{\text{unit}}(v) = ||\omega||_2 - 1
\]

Here, \(\omega\) represents the learnable vector representing a node, and \(||\omega||_2\) is the L2 norm of the vector. The goal is to minimize this loss, which effectively encourages the learnable vectors to have a fixed length of 1. This helps in finding meaningful and unique values for the vector while accounting for different scaling factors.\\

Furthermore, using TopK Pooling loss ensures that top K selected indicative ROIs should have significantly different scores than those of the unselected nodes. It helps in identifying the most relevant ROIs. 
In order to achieve the desired behavior in our model, where selected nodes should exhibit scores close to 1, while unselected nodes should approach a score of 0, we employ a training strategy as in (\cite{Li2021}). This strategy involves ranking the sigmoid-transformed scores, $\tilde{s}^{(l)}_m$, in descending order for each instance $m$, resulting in $\hat{s}^{(l)}_m = [\hat{s}^{(l)}_{m,1}, \ldots, \hat{s}^{(l)}_{m,N^{(l)}}]$. To introduce diversity in these ranked scores across all $M$ training instances, we employ a specific constraint. In practice, we define the TPK loss using binary cross-entropy, given by:

\begin{align}
L^{(l)}_{\text{TPK}} &= -\frac{1}{M}\sum_{m=1}^{M}\left(\frac{1}{N^{(l)}}\sum_{i=1}^{k}\log(\hat{s}^{(l)}_{m,i}) + \sum_{i=1}^{N^{(l)}-k}\log(1 - \hat{s}^{(l)}_{m,i+k})\right) \label{eq:tpk-loss}
\end{align}

This loss encourages the scores of selected ROIs to be higher ensuring that they stand out from the rest of the nodes.\\

The total loss $L_{\text{total}}$ is a combination of these individual loss components:

\[
L_{\text{total}} = L_{\text{smoothL1}} + \sum_{l=1}^{L} L^{(l)}_{\text{unit}} + \lambda_1 \sum_{l=1}^{L} L^{(l)}_{\text{TPK}}
\]

Where \(L\) represents the number of layers in the model, and \(\lambda_1\) is a hyperparameter that balances the contribution of the TopK Pooling loss. This combination of loss components helps in training the model to achieve good convergence, meaningful node representations, and relevant ROIs selection.\\

Overall with the architecture above, our proposed model for resting state fMRI data prediction using staticFNC could bring several benefits:\\

\begin{itemize}
\item 
By replacing the aggregation function of GIN with that of RGCN, the resulting RGIN model takes advantage of both  permutation invariant update function and aggregation functions.\\

\item We tackle the limitation of all brain ROIs learning embeddings in an identical manner by incorporating a clustering-based embedding method in the convolutional layer of the graph.\\

\item Instead of using conventional READOUT methods of GNNs, we make use of attention-based readout function which effectively captures the information by obtaining the importance scores for each node.

\end{itemize}

\section{Experiments}

\subsubsection{The ABCD (Adolescent Brain Cognitive Development) Dataset}

ABCD is a large ongoing study following youths from age 9-10 into late adolescence to understand factors that increase
the risk of physical and mental health problems. Participants were recruited from 21 sites across the US to represent various demographic variables. Data used in this study were from 8520 children aged 9–10 at baseline, including resting state fMRI image and fluid intelligence, crystallized intelligence, and total composite scores. Data were split into training (n = 5964), validation (n = 1278), and test (n = 1278) subsets. There were 4430 male and 4089 female subjects in this study. We note that age and site effects were regressed out from the intelligence scores to make sure that only the relevant intelligence features are captured by our graph model.\\

We conducted preprocessing on the raw resting-state fMRI data utilizing a combination of the FMRIB Software Library (FSL) v6.0 toolbox and Statistical Parametric Mapping (SPM) 12 toolbox within the MATLAB 2019b environment. The preprocessing encompassed several key steps, namely: 1) correction for rigid body motion; 2) distortion correction; 3) removal of dummy scans; 4) standardization to the Montreal Neurological Institute (MNI) space; and 5) application of a 6 mm Gaussian kernel for smoothing. Subsequently, we employed a fully automated spatially constrained independent component analysis framework to extract 53 robust intrinsic connectivity networks (ICNs) using the Neuromark\_fmri\_1.0 template. These ICNs were categorized into seven functional domains based on their anatomical locations and functional characteristics, which included subcortical, auditory, visual, sensorimotor, cognitive control, default mode, and cerebellar domains. Functional network connectivity (FNC) was computed as the Pearson correlation between the time courses of these intrinsic connectivity networks.

\begin{table*}[]
\caption{Comparison of different \technique  architectures for intelligence prediction}
\label{tab:graph-models-compare}
\resizebox{\linewidth}{!}{%
\begin{tabular}{|l|l|ll|ll|ll|}
\hline
\multicolumn{1}{|c|}{\multirow{2}{*}{\textbf{Aggregation}}} & \multicolumn{1}{c|}{\multirow{2}{*}{\textbf{Readout}}} & \multicolumn{2}{c|}{\textbf{Fluid Intelligence}} & \multicolumn{2}{c|}{\textbf{Crystallized Intelligence}} & \multicolumn{2}{c|}{\textbf{Total Composite Scores}} \\ \cline{3-8} 
\multicolumn{1}{|c|}{} & \multicolumn{1}{c|}{} & \multicolumn{1}{l|}{MSE} & Corr & \multicolumn{1}{l|}{MSE} & Corr & \multicolumn{1}{l|}{MSE} & Corr \\ \hline
Sum & Attention Based Readout (SERO) & \multicolumn{1}{l|}{\textbf{263}} & \textbf{0.23} & \multicolumn{1}{l|}{295.95} & 0.27 & \multicolumn{1}{l|}{\textbf{261}} & \textbf{0.31} \\ \cline{1-1} \cline{3-8} 
Mean &  & \multicolumn{1}{l|}{269} & 0.21 & \multicolumn{1}{l|}{295} & 0.27 & \multicolumn{1}{l|}{268.58} & 0.29  \\ \hline
Sum & Attention Based Readout (GARO) & \multicolumn{1}{l|}{274.69} & 0.20 & \multicolumn{1}{l|}{\textbf{263.7}} & \textbf{0.30} & \multicolumn{1}{l|}{291} & 0.29 \\ \cline{1-1} \cline{3-8} 
Mean &  & \multicolumn{1}{l|}{270.05} & 0.20 & \multicolumn{1}{l|}{268.7} & 0.28 & \multicolumn{1}{l|}{267.01} & 0.29 \\ \hline
Sum & \multirow{2}{*}{Mean + Max Readout} & \multicolumn{1}{l|}{300} & 0.19 & \multicolumn{1}{l|}{299.81} & 0.27 & \multicolumn{1}{l|}{271.37} & 0.30 \\ \cline{1-1} \cline{3-8} 
Mean &  & \multicolumn{1}{l|}{298} & 0.19 & \multicolumn{1}{l|}{298.30} & 0.27 & \multicolumn{1}{l|}{289.93} & 0.30 \\ \hline
\end{tabular}%
}
\end{table*}

\subsection{Experimental Setup}

The neural network architecture depicted in Fig. \ref{BrainRGIN_Architecture}  was implemented using Pytorch (\cite{PazkeTorch}), and Pytorch Geometric (\cite{TorchGeometric}) for the specific graph neural network components. The number of nodes was 53 (corresponding to NeuroMark (\cite{NeuroMark}) ICA Components), and the number of node features per each node was the number of pairwise correlations between each other nodes (i.e., 53). The performance of the \technique architecture was evaluated using the ABCD dataset for predicting fluid intelligence, crystallized intelligence, and total composite scores. We created three different variations of \technique  for various intelligence prediction tasks. The number of edges depends on the threshold percentage used to retain only the strongest correlations, which was selected to be 100\% for all the prediction tasks. The model architecture for all intelligence prediction tasks was implemented with 3 RGIN layers with node sizes 32, 128, and 256 respectively followed by pooling layers as shown in Fig. \ref{BrainRGIN_Architecture} and attention-based readout functions for each layer. The pooling ratio from hyperparameter tuning was set to be 0.38 from fluid intelligence, 0.46 for crystallized intelligence and 0.78 for total composite scores. The number of clustered communities is set to 7, and the motivation for this comes from the seven functional networks defined by (\cite{ThomasYeo2011}), because these 7 networks show key brain functionality relevance. The majority of these settings were fixed with the hyperparameter search. We also tested the \technique architecture with a variety of aggregation (Sum, Mean) and readout (SERO, GARO and Mean+Max) architectures to identify how different choices affect the performance of the model. We first identified the best \technique architecture compared with a variety of graph convolution, pooling, and readout architectures for predicting intelligence. Finally, the best architecture was used for comparison with baseline models.\\\\
We compared the performance of \technique with popular recent models along with traditional robust ML models as baseline models. The recent baselines includes BrainGNN (\cite{Li2021}), BrainNetCNN (\cite{BrainNetCNN}), Brain Network Transformer \cite{BrainNetworkTransformer}, FBNetGen \cite{FBNetGen}, Graph Transformers \cite{GraphTransformer} etc. We also made comparisons with traditional established ML methods such as Support Vector Regression (SVR), Logistic Regression (LR), and Ridge Regression (Ridge). We particularly excluded reporting the performance of other baseline models like GCN (\cite{GCN2017}), GAT (\cite{GAT}), GraphSage (\cite{GraphSage}) because of proven BrainGNN superior performance over these models. For BrainGNN, we performed multiple sweeps to identify the best performance. Based on performance, setting the learning rate of both BrainRGIN and BrainGNN initially  to 0.001 and reducing every 30 epochs achieved the best performance for all prediction tasks. For BrainGNN, we used two convolutional layers of size 32, as presented in the original paper (\cite{Li2021}), with a fully connected layer of size 512. We utilized the authors' provided open-source codes for BrainGNN \cite{Li2021}, FBNetGen \cite{FBNetGen} and Brain Network Transformer (BNT) \cite{BrainNetworkTransformer}, conducting grid searches to fine-tune essential hyperparameters based on the recommended best configurations. Specifically, for BrainGNN, we also explored various learning rates (0.01, 0.005, 0.001) while for FBNetGen, we investigated how varying hidden dimensions (8, 12, 16) impacts the performance. For BNT, we used different hidden dimensions (256, 512, 1024) during scaled dot product attention, along with varying fully connected layers (2 and 3).  In the case of BrainNetCNN \cite{BrainNetCNN}, we experimented with different dropout rates (0.3, 0.5, 0.7). Regarding GT (Graph Transformer) \cite{GraphTransformer}, we examined the impact of different number of transformer layers (1, 2, 4) and the number of attention heads (2, 4, 8) over 50 epochs of training. For the remaining baseline methods, we employed Polyssifier, a widely used tool for baseline model comparisons, leveraging its automatic hyperparameter search feature to identify optimal model parameters for maximizing performance. For the other traditional baseline methods, we used Polyssifier (https://github.com/sergeyplis/polyssifier). This tool is widely used to perform baseline model comparisons and has a feature to automatically perform hyperparameter search and identify the best parameters of the model for the highest performance. For the evaluation metrics, we considered Mean Squared Error (MSE) and Correlation.  The results were reported using a separate holdout testing set (n=1278) and the average scores of experiments using four different seeds are reported for every intelligence prediction task.

\section{Results}

The experimental results are summarized in Table \ref{tab:graph-models-compare} and Table \ref{tab:baseline-models-compare}. In Table \ref{tab:graph-models-compare}, it is evident that the \technique architecture demonstrated promising performance in predicting fluid intelligence, achieving a mean squared error (MSE) of 263 and a correlation coefficient of 0.23 when employing RGIN convolution combined with the SERO readout method. Moreover, it yielded the best results in predicting crystallized intelligence, with an MSE of 263.7 and a correlation of 0.30 using the same RGIN convolution and SERO readout method. Notably, for total composite scores, the GARO attention-based readout function in conjunction with the RGIN graph model attained the highest performance, achieving an MSE of 261 and a correlation of 0.31.\\\\
Furthermore, we observed stable and comparable performances from baseline models such as BrainNetCNN \cite{BrainNetCNN}, FBNetGen \cite{FBNetGen}, and GT \cite{GraphTransformer}, indicating the reliability of these models. However, the Brain Network Transformer (BNT) \cite{BrainNetworkTransformer} exhibited robust performance with higher correlation scores and lower mean squared errors. Although \technique occasionally reported lower mean squared errors compared to BNT, the performance appeared to be comparable. Additionally, \technique surpassed all traditional baseline models and exhibited superior performance compared to Support Vector Regression (SVR), Linear Regression (LR), and Ridge Regression. It consistently achieved lower MSE values and higher correlation coefficients across all metrics, namely fluid intelligence, crystallized intelligence, and total composite scores. These results underscore the effectiveness of RGIN aggregation and attention-based readout methods over the BrainGNN model and other baseline models in predicting intelligence scores. Moreover, attention-based readout methods were found to outperform other readout techniques, significantly contributing to the architecture's success in predicting intelligence scores.

\begin{table*}[]
\caption{Comparison of \technique With Baseline Models on ABCD Dataset}
\label{tab:baseline-models-compare}
\resizebox{\textwidth}{!}{%
\begin{tabular}{|lllllllll|}
\hline
\multicolumn{1}{|l|}{} & \multicolumn{1}{l|}{} & \multicolumn{2}{c|}{\textbf{Fluid Intelligence}} & \multicolumn{2}{c|}{\textbf{Crystallized Intelligence}} & \multicolumn{2}{c|}{\textbf{Total Composite Scores}} &  \\ \cline{3-9} 
\multicolumn{1}{|l|}{\multirow{-2}{*}{\textbf{Models}}} & \multicolumn{1}{l|}{\multirow{-2}{*}{\textbf{Linear}}} & \multicolumn{1}{l|}{MSE} & \multicolumn{1}{l|}{Correlation} & \multicolumn{1}{l|}{MSE} & \multicolumn{1}{l|}{Correlation} & \multicolumn{1}{l|}{MSE} & \multicolumn{1}{l|}{Correlation} &  \\ \hline
\multicolumn{1}{|l|}{SVR} & \multicolumn{1}{l|}{No} & \multicolumn{1}{c|}{275} & \multicolumn{1}{l|}{0.20} & \multicolumn{1}{c|}{307} & \multicolumn{1}{l|}{0.24} & \multicolumn{1}{c|}{307} & \multicolumn{1}{l|}{0.24} &  \\ \hline
\multicolumn{1}{|l|}{LR} & \multicolumn{1}{l|}{Yes} & \multicolumn{1}{c|}{326} & \multicolumn{1}{l|}{0.19} & \multicolumn{1}{c|}{354} & \multicolumn{1}{l|}{0.22} & \multicolumn{1}{c|}{345} & \multicolumn{1}{l|}{0.23} &  \\ \hline
\multicolumn{1}{|l|}{Ridge} & \multicolumn{1}{l|}{Yes} & \multicolumn{1}{c|}{325} & \multicolumn{1}{l|}{0.19} & \multicolumn{1}{c|}{353} & \multicolumn{1}{l|}{0.22} & \multicolumn{1}{c|}{349} & \multicolumn{1}{l|}{0.23} &  \\ \hline
\multicolumn{1}{|l|}{BrainGNN \cite{Li2021}} & \multicolumn{1}{l|}{No} & \multicolumn{1}{l|}{264} & \multicolumn{1}{l|}{0.22} & \multicolumn{1}{l|}{288.29} & \multicolumn{1}{l|}{0.28} & \multicolumn{1}{l|}{294.12} & \multicolumn{1}{l|}{0.29} &  \\ \hline
\multicolumn{1}{|l|}{\technique (Ours)} & \multicolumn{1}{l|}{No} & \multicolumn{1}{l|}{263} & \multicolumn{1}{l|}{0.23} & \multicolumn{1}{l|}{263.7} & \multicolumn{1}{l|}{0.30} & \multicolumn{1}{l|}{261} & \multicolumn{1}{l|}{0.31} &  \\ \hline
\multicolumn{1}{|l|}{{\color[HTML]{000000} BrainNetCNN \cite{BrainNetCNN}}} & \multicolumn{1}{l|}{{\color[HTML]{000000} No}} & \multicolumn{1}{l|}{{\color[HTML]{000000} 372.32}} & \multicolumn{1}{l|}{{\color[HTML]{000000} 0.17}} & \multicolumn{1}{l|}{{\color[HTML]{000000} 367.82}} & \multicolumn{1}{l|}{{\color[HTML]{000000} 0.204}} & \multicolumn{1}{l|}{{\color[HTML]{000000} 375.26}} & \multicolumn{1}{l|}{{\color[HTML]{000000} 0.218}} &  \\ \hline
\multicolumn{1}{|l|}{{\color[HTML]{000000} Brain Network Transformer \cite{BrainNetworkTransformer}}} & \multicolumn{1}{l|}{{\color[HTML]{000000} No}} & \multicolumn{1}{l|}{{\color[HTML]{000000} 261.59}} & \multicolumn{1}{l|}{{\color[HTML]{000000} 0.248}} & \multicolumn{1}{l|}{{\color[HTML]{000000} 265.12}} & \multicolumn{1}{l|}{{\color[HTML]{000000} 0.32}} & \multicolumn{1}{l|}{{\color[HTML]{000000} 264.65}} & \multicolumn{1}{l|}{{\color[HTML]{000000} 0.339}} & {\color[HTML]{000000} } \\ \hline
\multicolumn{1}{|l|}{{\color[HTML]{000000} FBNetGen \cite{FBNetGen}}} & \multicolumn{1}{l|}{{\color[HTML]{000000} No}} & \multicolumn{1}{l|}{{\color[HTML]{000000} 265.21}} & \multicolumn{1}{l|}{{\color[HTML]{000000} 0.21}} & \multicolumn{1}{l|}{{\color[HTML]{000000} 276.27}} & \multicolumn{1}{l|}{{\color[HTML]{000000} 0.315}} & \multicolumn{1}{l|}{{\color[HTML]{000000} 273.74}} & \multicolumn{1}{l|}{{\color[HTML]{000000} 0.30}} & {\color[HTML]{000000} } \\ \hline
\multicolumn{1}{|l|}{{\color[HTML]{000000} Graph Transformers \cite{GraphTransformer}}} & \multicolumn{1}{l|}{{\color[HTML]{000000} No}} & \multicolumn{1}{l|}{{\color[HTML]{000000} 267}} & \multicolumn{1}{l|}{{\color[HTML]{000000} 0.22}} & \multicolumn{1}{l|}{{\color[HTML]{000000} 281}} & \multicolumn{1}{l|}{{\color[HTML]{000000} 0.23}} & \multicolumn{1}{l|}{{\color[HTML]{000000} 278}} & \multicolumn{1}{l|}{{\color[HTML]{000000} 0.32}} & {\color[HTML]{000000} } \\ \hline
\end{tabular}%
}
\end{table*}

\begin{table*}[t]
\caption{Effect of edges threshold selection in model prediction}
\centering
\label{tab:edges-hyperparams}
\resizebox{0.6\linewidth}{!}{%
\begin{tabular}{|l|lll|}
\hline
\multirow{2}{*}{\textbf{Edge Threshold}} & \multicolumn{3}{c|}{\textbf{MSE}} \\ \cline{2-4} 
 & \multicolumn{1}{l|}{\textbf{Fluid Intelligence}} & \multicolumn{1}{l|}{\textbf{Crystallized Intelligence}} & \textbf{Total Composite Intelligence} \\ \hline
 100\% & \multicolumn{1}{l|}{263} & \multicolumn{1}{l|}{263.7} & 261 \\ \hline
 80\% & \multicolumn{1}{l|}{269.22} & \multicolumn{1}{l|}{276.12} & 272 \\ \hline
60\% & \multicolumn{1}{l|}{271.35} & \multicolumn{1}{l|}{278.38} & 269.24 \\ \hline
50\% & \multicolumn{1}{l|}{288.12} & \multicolumn{1}{l|}{281.06} & 283 \\ \hline
40\% & \multicolumn{1}{l|}{290.05} & \multicolumn{1}{l|}{297.12} & 288 \\ \hline
\end{tabular}%
}
\end{table*}

 These findings underscore the importance of carefully selecting appropriate graph neural network components for predicting intelligence scores from rsfMRI data and provide a valuable foundation for future research in this area.
\subsection{Hyperparameter Study}
We also analyzed how changing the edges threshold affected the performance of the model. Interestingly, as can be seen from Table \ref{tab:edges-hyperparams}, we saw that using 100\% of edges gave a superior performance for prediction for all intelligence prediction tasks. 
Besides, inspired by BrainGNN (\cite{Li2021}) hyperparameter study for TopK Loss regularizer, we fixed the regularizer $\lambda1$ to 0.1 to partially force the network to select the same top nodes and also make sure that the selected nodes have different scores than those of the unselected nodes.

\subsection{Interpretation of brain regions and networks}

\begin{figure}[t]
    \centering
    \begin{subfigure}{0.45\linewidth}
        \includegraphics[width=\linewidth]{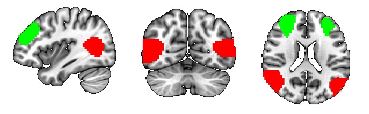}
        \caption{ Middle frontal gyrus and Middle temporal gyrus}
        \label{fig:fluid}
    \end{subfigure}
    \hfill
    \begin{subfigure}{0.45\linewidth}
        \includegraphics[width=\linewidth]{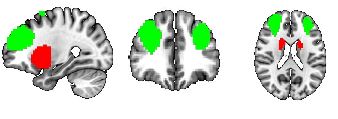}
        \caption{Middle frontal gyrus and Caudate}
        \label{fig:crystallized}
    \end{subfigure}
    \\
    \caption{Regions Significant in Fluid  and Crystallized Intelligence Prediction}
    \label{fig:fluid_intelligence_regions}
\end{figure}

To identify the top ROIs for the intelligence prediction tasks, we first extracted the pooled ROIs after passing through the first layer of TopK Pool layer of a fully trained model using a holdout test set. We then selected the ROIs with highest frequency across each of the samples for all the intelligence scores. For fluid intelligence, we selected two ROIs (Middle Temporal Gyrus, and Middle Frontal Gyrus) (Fig \ref{fig:fluid_intelligence_regions}(a)) with 90\% frequency threshold. Similarly, for crystallized intelligence, middle frontal gyrus and caudate were selected with a frequency threshold of 90\% (Fig \ref{fig:fluid_intelligence_regions}(b)). For total composite scores, as the pooling ratio from hyper-parameter tuning was higher (0.78), we were able to extract 21 brain regions for 95\% frequency. We categorized the selected regions into seven connectivity networks and plotted them independently in Fig. 3. \\

\subsection{Experiment with HCP dataset}
\subsubsection{HCP Dataset}
For generalizability, we used HCP \cite{VanEssen2013} data was first minimally pre-processed following the pipeline described in \cite{Glasser2013}. The preprocessing includes gradient distortion correction, motion correction, and field map preprocessing, followed by registration to T1 weighted images. The registered EPI image was then normalized to the standard MNI152 space. To reduce noise from the data, FIX-ICA based denoising was applied \cite{ SalimiKhorshidi2014}. To minimize the effects of head motion subject scans with framewise displacement (FD) over 0.3mm at any time of the scan were discarded. The FD was computed with fsl motion outliers function of the FSL \cite{Jenkinson2012}. There were 152 discarded scans from filtering out with the FD, and 1075 scans were left. For all experiments, the scans from the first run of HCP subjects released under S1200 were used. Subsequently, similar to preprocessing with ABCD dataset, we employed a fully automated spatially constrained independent component analysis framework to extract 53 robust intrinsic connectivity networks (ICNs) using the Neuromark fmri 1.0 template which categorized into seven functional domains based on
their anatomical locations and functional characteristics. Furthermore, we also regressed out age and gender effects from the intelligence scores to make sure the model captures relevant feature representations.

\subsubsection{Hyperparameter settings for HCP}
As the measures of intelligence on the HCP dataset are different than those of ABCD, we retrained the model with different hyperparameters as we expected configuration changes as compared to the ABCD architecture. Similar to the ABCD dataset, we created three variations of \technique for various intelligence prediction tasks. The edges threshold percentage was set to 100\% referencing the earlier ABCD results. Similar to ABCD experiments, the model was trained with 3 RGIN layers with node sizes 32, 128, and 256 followed by a pooling layer and attention-based readout as shown in Fig \ref{BrainRGIN_Architecture}. The pooling ratio from sweeps from hyperparameter tuning was set to be 0.42 for fluid, 0.52 for crystallized, and 0.65 for total composite scores. We used sum aggregation with SERO readout which was found to be robust with ABCD experiments. The results reported in Table \ref{tab:baseline-models-compare-hcp} are the average scores of experiments using four different seeds for every intelligence prediction task on separate holdout testing sets (n=192). \\

\subsubsection{Results from HCP dataset}
The performance of various models on the HCP dataset is presented in Table \ref{tab:baseline-models-compare-hcp}. Notably, our model, \technique, demonstrates competitive performance across all intelligence scores. For fluid intelligence prediction, BrainRGIN achieves a correlation coefficient of 0.262 with a mean squared error (MSE) of 274.92. Similarly, for predicting crystallized intelligence, BrainRGIN achieves a correlation coefficient of 0.2937 with an MSE of 308.21. Additionally, for total composite scores, BrainRGIN achieves a correlation coefficient of 0.309 with an MSE of 306.45. Comparatively, Brain Network Transformer and FBNetGen also exhibit noteworthy performance, achieving very comparable results. However, similar to ABCD experiments, our model reports slightly lower mean squared errors in some cases, indicating its potential for more precise predictions based on a graph convolution method.\\

\begin{table*}[]
\caption{Comparison of \technique With Baseline Models on HCP Dataset}
\label{tab:baseline-models-compare-hcp}
\resizebox{\textwidth}{!}{%
\begin{tabular}{|l|ll|ll|ll|l|}
\hline
{\color{black} \textbf{Model}} & \multicolumn{2}{c|}{\textbf{\textcolor{black}{Fluid}}}             & \multicolumn{2}{c|}{\textbf{\textcolor{black}{Crystallized}}} & \multicolumn{2}{c|}{\textbf{\textcolor{black}{TotalComp}}} & {\color{black} } \\ \hline
\multirow{-2}{*}{{\color{black} }} & \multicolumn{1}{l|}{\textbf{\textcolor{black}{Correlation}}} & {\color{black} \textbf{MSE}} & \multicolumn{1}{l|}{\textbf{\textcolor{black}{Correlation}}} & {\color{black} \textbf{MSE}} & \multicolumn{1}{l|}{\textbf{\textcolor{black}{Correlation}}} & {\color{black} \textbf{MSE}} & {\color{black} } \\ \hline
{\color{black} \technique (Ours)} & \multicolumn{1}{l|}{\textcolor{black}{0.262}} & \textcolor{black}{274.92} & \multicolumn{1}{l|}{\textcolor{black}{0.2937}} & \textcolor{black}{308.21} & \multicolumn{1}{l|}{\textcolor{black}{0.309}} & \textcolor{black}{306.45} & {\color{black} } \\ \hline
{\color{black} BrainGNN \cite{Li2021}} & \multicolumn{1}{l|}{\textcolor{black}{0.256}} & \textcolor{black}{278.81} & \multicolumn{1}{l|}{\textcolor{black}{0.285}} & \textcolor{black}{312.52} & \multicolumn{1}{l|}{\textcolor{black}{0.285}} & \textcolor{black}{314.233} & {\color{black} } \\ \hline
{\color{black} Brain Network Transformer \cite{BrainNetworkTransformer}} & \multicolumn{1}{l|}{\textcolor{black}{0.271}} & \textcolor{black}{278.51} & \multicolumn{1}{l|}{\textcolor{black}{0.302}} & \textcolor{black}{303.26} & \multicolumn{1}{l|}{\textcolor{black}{0.318}} & \textcolor{black}{312.44} & {\color{black} } \\ \hline
{\color{black} BrainNetCNN \cite{BrainNetCNN}} & \multicolumn{1}{l|}{\textcolor{black}{0.251}} & \textcolor{black}{303.21} & \multicolumn{1}{l|}{\textcolor{black}{0.268}} & \textcolor{black}{328.14} & \multicolumn{1}{l|}{\textcolor{black}{0.271}} & \textcolor{black}{322.23} & {\color{black} } \\ \hline
{\color{black} FBNetGen \cite{FBNetGen}} & \multicolumn{1}{l|}{\textcolor{black}{0.257}} & \textcolor{black}{281.21} & \multicolumn{1}{l|}{\textcolor{black}{0.291}} & \textcolor{black}{307.38} & \multicolumn{1}{l|}{\textcolor{black}{0.301}} & \textcolor{black}{316.34} & {\color{black} } \\ \hline
{\color{black} Graph Transformer \cite{GraphTransformer}} & \multicolumn{1}{l|}{\textcolor{black}{0.256 }} & \textcolor{black}{361} & \multicolumn{1}{l|}{\textcolor{black}{0.28}} & \textcolor{black}{318} & \multicolumn{1}{l|}{\textcolor{black}{0.267}} & \textcolor{black}{372.32} & {\color{black} } \\ \hline
\end{tabular}%
}
\end{table*}

\section{Discussion and Conclusion}

In this research study, a novel technique called Brain ROI-Aware Graph Isomorphism Networks, \technique, was proposed to predict intelligence using static FNC matrices derived from resting-state fMRI data. \technique integrates the expressive power of GIN and clustering-based GCN and also incorporates attention-based readout function, in the hope of better representing brain networks and improving model prediction. Specifically, by replacing the aggregation function of GIN with that of RGCN, the model leverages the powerful representation learning capability of GIN, while still capturing the edge strength and edge type represented by a clustering-based embedding method. Another notable aspect of the proposed architecture is the use of attention-based readout functions instead of conventional readout methods, which is proven to be very effective as can be seen from Table \ref{tab:graph-models-compare}. The attention mechanism assigns importance scores to each node, effectively capturing spatial relevance information for prediction. Using attention-based readout functions not only improved the overall prediction of the model but also validated the theory that different brain regions contribute in a different manner to intelligence prediction.\\

\begin{figure}
    \centering
    
    \begin{subfigure}{0.45\linewidth}
        \includegraphics[width=\linewidth]{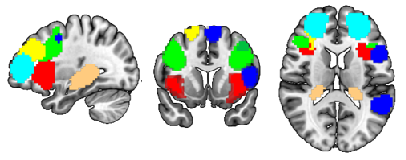}
        \caption{Cognitive Control Network}
        \label{fig:cognitive}
    \end{subfigure}
    \hfill
    \begin{subfigure}{0.45\linewidth}
        \includegraphics[width=\linewidth]{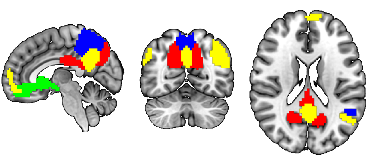}
        \caption{Default Mode Network}
        \label{fig:defaultmode}
    \end{subfigure}
    
    \vspace{0.5cm} 
    
    \begin{subfigure}{0.45\linewidth}
        \includegraphics[width=\linewidth]{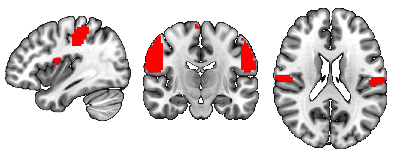}
        \caption{Sensorimotor Network}
        \label{fig:sensorimotor}
    \end{subfigure}
    \hfill
    \begin{subfigure}{0.45\linewidth}
        \includegraphics[width=\linewidth]{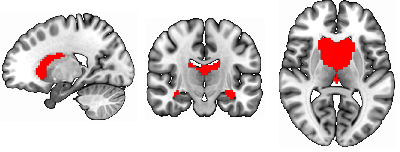}
        \caption{Subcortical Network}
        \label{fig:subcortical}
    \end{subfigure}
    
    \vspace{0.5cm} 
    
    \begin{subfigure}{0.45\linewidth}
        \includegraphics[width=\linewidth]{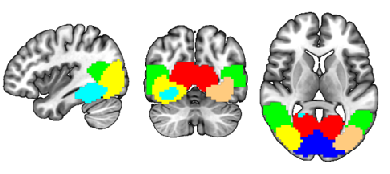}
        \caption{Visual Network}
        \label{fig:visual}
    \end{subfigure}
    \hfill
    \begin{subfigure}{0.45\linewidth}
        \includegraphics[width=\linewidth]{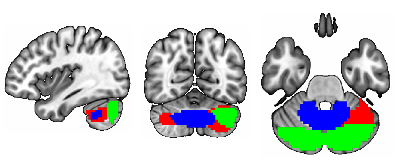}
        \caption{Cerebellar Networks}
        \label{fig:cerebellar}
    \end{subfigure}
    
    \caption{Significant regions expressed as Connectivity Networks for Total Composite Scores}
    \label{fig:six_images}
\end{figure}

To ensure the model is robust to influence from demographic factors such as age, gender, site, and socioeconomic status, we have addressed demographic factors such as age, gender, and site (which is expected to have a huge effect on MRI data) being controlled for the experiment. Specifically, we effectively regressed these factors from intelligence scores during model evaluation to mitigate their influence. It's noteworthy that before regressing out site from intelligence scores, the final correlation scores were as high as 0.35 for fluid intelligence, 0.42 for crystallized intelligence, and 0.41 for total composite scores with lower mean squared errors. However, removing these covariates led to a drop in overall prediction scores based on correlation and MSE metrics as presented in Table \ref{tab:baseline-models-compare}. Furthermore, we did an in-depth analysis to assess the potential impact of socioeconomic status (SES), focusing on income and education, on our model evaluation for crystallized intelligence. Originally, our data revealed moderate positive correlations between SES factors and predicted (0.28) and original crystallized intelligence scores (0.33). To further evaluate the influence of SES, we retrained the model on crystallized intelligence, controlling for income and education to generate new intelligence scores, keeping the model configuration consistent with earlier training. Surprisingly, upon removing SES factors from the evaluation, the model's performance improved slightly, as evidenced by an increased correlation of 0.331 from 0.30 on crystallized intelligence and a lower mean squared error of 256.21 from 263.7. The top regions of interest were similar to the earlier model where we did not regress out SES. These findings robustly demonstrate that our model evaluation remains unaffected by socioeconomic status and other demographic factors.\\


The top two regions contributing to fluid intelligence are the middle frontal gyrus and the middle temporal gyrus. In fact middle frontal gyrus contributes to all three intelligence scores, and its overall significance in predicting intelligence is in line with the literature. In particular, the dorsolateral prefrontal cortex in the middle frontal gyrus is well known to play a crucial part in many domains of cognitive processes, including working memory (\cite{Barbey2013}), attention control (\cite{Knight2020}), executive function (\cite{Nejati2021}), and decision making (\cite{Patterson2007}), etc (\cite{Yu2022}). Given its multi-domain involvement in cognition, this region's engagement substantiates its relevance in all intelligence scores. The function of the middle temporal gyrus is relatively more specific to language relevant (\cite{Turken2011}), such as semantic processing (\cite{Binder2009}), syntactic comprehension (\cite{Yu2022}), language comprehension, decoding intelligible speech. Its role in fluid intelligence prediction could be attributed to its contribution to information decoding and  integration. In the context of crystallized intelligence, besides the middle frontal gyrus, caudate plays the most remarked contribution. The caudate's association with learning (\cite{Choi2020}) and memory consolidation (\cite{Mller2017}) aligns with crystallized intelligence's reliance on acquired knowledge and experience (\cite{Grazioplene2014}). The combined involvement of middle frontal gyrus and caudate suggests a cooperative engagement in translating stored knowledge into practical problem-solving skills, a hallmark of crystallized intelligence. 

The diverse set of brain regions identified as relevant for total composite score reflects the intricate and distributed nature of cognitive processes, as well as the general and broad attribute of 'intelligence'. These regions cover six relatively separate brain networks. Sensorimotor network (postcentral gyrus) and visual network(middle/inferior occipital gyri, calcarine cortex, cuneus, fusiform) involve basic functions of various sensory inputs (\cite{Postcentral}) and further integration of inputs (\cite{BrainMadeSimple_2022}). The cognitive control network covers particularly frontal-parietal regions (superior/middle frontal gyri, supplementary motor area, insula, hippocampal) and participates in various higher-order cognitive processes (\cite{Jacques2018}), such as executive function, attention, working memory, and planning (\cite{Tanji1994}). The default network comprises the anterior cingulate cortex, posterior cingulate cortex, and precuneus and reflects the brain's intrinsic activity in contrast to task-oriented controls (\cite{Zanto2013}). Their functions are linked to self-referential thinking (\cite{Raichle2015}), introspection, and episodic memory retrieval (\cite{Gusnard2001}). Subcortical network, precisely caudate as discussed before has a role in learning and memory and has been shown to be able to predict IQ (\cite{Vincent2006}). Last but not the least is the cerebellum. Its function is well recognized to be beyond motor function, but subserving cognition (\cite{Koziol2011}), and recently cerebellum has been partitioned to echo the 7 cortical networks (\cite{King2019}).  
In summary, the interplay of these diverse brain regions likely contributes to the complexity of cognitive processes that manifest as Total Composite Scores. The extensive network of regions suggests that intelligence is not confined to a single brain area but rather emerges from the collaboration of numerous interconnected regions, each contributing its unique specialization to overall cognitive performance.

In conclusion, the \technique architecture presented in this study offers several advantages for intelligence prediction using resting-state fMRI data. It effectively captures the local and global relationships between brain regions, incorporates edge information, utilizes attention-based readout functions, and addresses the variability among brain ROIs. The superior performance of \technique over baseline models highlights its potential for predicting intelligence and provides valuable insights into the relationship between brain function and cognitive processes. Future research can explore the application of this architecture to other types of fMRI data, such as task-based fMRI, to further advance our understanding of brain functioning.\\

\section*{Acknowledgements}

This study was funded part by NIH grant R01DA049238 and NSF grant 2112455. We thank the Adolescent Brain Cognitive Development (ABCD) participants and their families for their time and dedication to this project. Data used in the preparation of this article were obtained from the Adolescent Brain Cognitive Development (ABCD) Study (https://abcdstudy.org), held in the NIMH Data Archive (NDA). This is a multisite, longitudinal study designed to recruit more than 10,000 children aged 9-10 and follow them over 10 years into early adulthood. The ABCD Study is supported by the National Institutes of Health and additional federal partners under award numbers U01DA041048, U01DA050989, U01DA051016, U01DA041022, U01DA051018, U01DA051037, U01DA050987, U01DA041174, U01DA041106, U01DA041117, U01DA041028, U01DA041134, U01DA050988, U01DA051039, U01DA041156,
U01DA041025, U01DA041120, U01DA051038, U01DA041148, U01DA041093, U01DA041089, U24DA041123,
U24DA041147. A full list of supporters is available at https://abcdstudy.org/federal-partners.html. A listing of participating sites and a complete listing of the study investigators can be found at https://abcdstudy.org/consortium\_members. ABCD consortium investigators designed and implemented the study and/or provided data but did not necessarily participate in analysis or writing of this report. This manuscript reflects the views of the authors and may not reflect the opinions or views of the NIH or ABCD consortium investigators.

\section*{Author contributions statement}

B.T. conducted the primary data analysis, coding implementation, experiments, writing, and revision of the manuscript. R.S, B.R., and P.S were involved in the in-depth discussion of the analyses and reviewing the manuscript. J.C. assisted in analyzing the results, and also provided assistance in interpretation. V.D.C was involved in the design of the project and the manuscript review.  J.L. was involved in the design, data analysis, and revision of the manuscript. E.A was involved in writing and editing the manuscript. All authors reviewed the manuscript. 

\section*{Code and Data Availability for Replication}

The code for replication is available at \url{https://github.com/bishalth01/BrainRGIN}. Data will be made available upon request.

\section*{Financial Disclosures Section}

All authors declare that they have no conflicts of interest.

\bibliography{main_manuscript} 
\bibliographystyle{unsrtnat}







\end{document}